\pdfoutput=1
\documentclass[runningheads]{llncs}

\usepackage{times}
\usepackage{latexsym}
\usepackage{wrapfig,lipsum,booktabs}
\usepackage{times}  
\usepackage{helvet} 
\usepackage{courier}  
\usepackage[hyphens]{url}  
\usepackage{graphicx} 
\usepackage[numbers]{natbib}  
\usepackage{caption}
\usepackage{mdframed}
\usepackage{flushend}
\RequirePackage{colortbl}
\definecolor{airforceblue}{rgb}{0.36, 0.54, 0.66}
\definecolor{amaranth}{rgb}{0.9, 0.17, 0.31}
\definecolor{applegreen}{rgb}{0.55, 0.71, 0.0}
\definecolor{alizarin}{rgb}{0.82, 0.1, 0.26}
\definecolor{azure}{rgb}{0.0, 0.5, 1.0}
\definecolor{cadmiumgreen}{rgb}{0.0, 0.42, 0.24}
\usepackage[T1]{fontenc}
\usepackage{latexsym}
\usepackage{url}
\usepackage{wrapfig}
\usepackage{lipsum,caption}
\usepackage{amsmath}
\usepackage{multirow}
\usepackage{graphicx}
\usepackage{color}
\usepackage[ruled,vlined]{algorithm2e}
\usepackage{xspace}
\usepackage{color,soul}
\usepackage[textsize=scriptsize]{todonotes}

\usepackage{physics}
\usepackage{tcolorbox}
\usepackage{xcolor}
\usepackage{amssymb}
\usepackage{comment}
\usepackage{booktabs}
\usepackage{longtable}
\usepackage{enumitem}
\usepackage{todonotes}
\RequirePackage{colortbl}
\usepackage{graphicx}
\usepackage[linguistics]{forest}
\usepackage[noend]{algorithmic}
\usepackage{listings}
\usepackage{times}  
\usepackage{helvet} 
\usepackage{courier}  
\usepackage[hyphens]{url}  
\usepackage{graphicx} 
\usepackage{graphicx}  
\usepackage{setspace}

\definecolor{dkgreen}{rgb}{0,0.6,0}
\definecolor{gray}{rgb}{0.5,0.5,0.5}
\definecolor{mauve}{rgb}{0.58,0,0.82}
\def\etal{\textit{et al.}}

\usepackage[T1]{fontenc}

\usepackage[utf8]{inputenc}

\usepackage{microtype}
\begin{document}
\title{On Adversarial Robustness of Synthetic Code Generation}
%
%
\author{Mrinal Anand\inst{1} \and
 Pratik Kayal\inst{1} \and
 Mayank Singh\inst{1}}
\authorrunning{Anand et al.}
%
\institute{$^1$Indian Institute of Technology Gandhinagar, India \\
\email{\{mrinal.anand,pratik.kayal,singh.mayank\}@iitgn.ac.}}
\maketitle              
\begin{abstract}
Automatic code synthesis from natural language descriptions is a challenging task. We witness massive progress in developing code generation systems for domain-specific languages (DSLs) employing sequence-to-sequence deep learning techniques in the recent past. In this paper, we specifically experiment with \textsc{AlgoLisp} DSL-based generative models and showcase the existence of significant dataset bias through different classes of adversarial examples. We also experiment with two variants of Transformer-based models that outperform all existing \textsc{AlgoLisp} DSL-based code generation baselines. Consistent with the current state-of-the-art systems, our proposed models, too, achieve poor performance under adversarial settings. Therefore, we propose several dataset augmentation techniques to reduce bias and showcase their efficacy using robust experimentation.

\end{abstract}
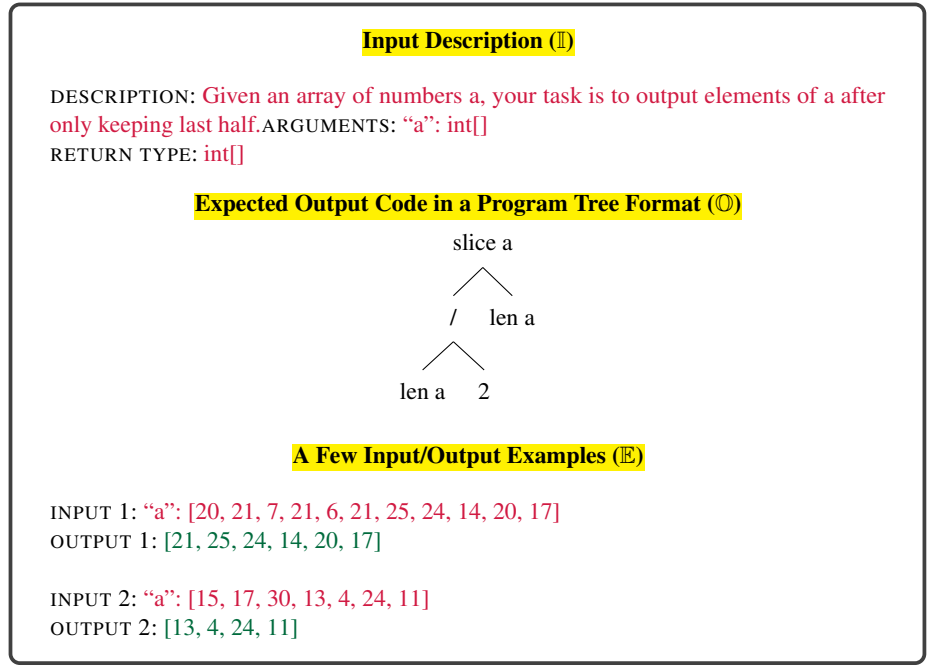
\begin{figure}[!b]
    \centering
    \begin{tcolorbox}[colback=white]
\small{
\begin{center}
\hl{\textbf{Input Description ($\mathbb{I}$)}}\\
\end{center}
\textsc{description}: \textcolor{alizarin}{Given an array of numbers a, your task is to output elements of a after only keeping last half.}
\textsc{arguments}: \textcolor{alizarin}{``a'': int[]}\\
\textsc{return type}: \textcolor{alizarin}{int[]}
\begin{center}
\hl{\textbf{Expected Output Code in a Program Tree Format ($\mathbb{O}$)}}\\
\begin{forest}
[slice a
 [/
    [len a]
    [2]
 ]
    [len a]
 ]
\end{forest}
\end{center}

\begin{center}
\hl{\textbf{A Few Input/Output Examples ($\mathbb{E}$)}}\\
\end{center}
\textsc{input 1}: \textcolor{alizarin}{``a'': [20, 21, 7, 21, 6, 21, 25, 24, 14, 20, 17]}\\
\textsc{output 1}: \textcolor{cadmiumgreen}{[21, 25, 24, 14, 20, 17]}\\

\textsc{input 2}: \textcolor{alizarin}{``a'': [15, 17, 30, 13, 4, 24, 11]}\\
\textsc{output 2}: \textcolor{cadmiumgreen}{[13, 4, 24, 11]}
}
\end{tcolorbox}
\caption{An illustrative example comprising a textual description ($\mathbb{I}$), its corresponding LISP program in a tree format ($\mathbb{O}$), and a few input-output examples ($\mathbb{E}$).}
    \label{fig:example_problem}
\end{figure}

\textit{Can computers automatically synthesize a program? Is it possible to generate code from a human-readable textual description?}  These are fundamental questions for any field of research, but particularly well-motivated in Computer Science. Automatic program synthesis (popularly known as \textit{`Code Generation'}) is the process of generating code in a particular language from a code/description in some other language. Figure~\ref{fig:example_problem} shows an illustrative example comprising a textual description, its corresponding LISP program in a tree format, and few input/output (I/O) pairs. Similar examples can be easily constructed from other popular programming languages like Python, Java, and C/C++, using different coding platforms\footnote{For, e.g., GeeksForGeeks (\url{https://www.geeksforgeeks.org})}.

\noindent\textbf{The Two Paradigms Of Code Synthesis} The rich literature on automatic code generation is broadly classified into two categories: (i) \textit{programming by example} (PBE)~\citep{menon2013machine}, and (ii) \textit{programming by descriptions} (PBD)~\cite{LinWPVZE2017:TR}. The PBE paradigm leverages input-output (I/O) examples ($\mathbb{E}$ in Figure~\ref{fig:example_problem}) alone to automatically construct a program ($\mathbb{O}$ in Figure~\ref{fig:example_problem}) that satisfies these examples. Several real-world computer science applications use the PBE paradigm for automatic code synthesis. For example, FlashFill~\cite{gulwani2011automating}, DeepCoder~\cite{balog2016deepcoder} and RobustFill~\cite{devlin2017robustfill}. 
In contrast, PBD paradigm uses descriptions ($\mathbb{I}$ in Figure~\ref{fig:example_problem}) with corresponding zero or few I/O code instances ($\mathbb{E}$ in Figure~\ref{fig:example_problem}) to automatically constructs a program ($\mathbb{O}$ in Figure~\ref{fig:example_problem}). The PBD paradigm has recently received major attention, thanks to the surge in the neural sequence-to-sequence approaches~\cite{LinWPVZE2017:TR,neuralprogramsearch2018,zavershynskyi2018naps}. However, the progress is fairly limited due to unavailability of large-scale real datasets. Polosukhin~\etal~\cite{neuralprogramsearch2018} proposed a large-scale synthetic dataset, \textsc{AlgoLisp}, and a corresponding neural architecture \textsc{Seq2Tree}. \textsc{Seq2Tree} generates \textit{Abstract Syntax Trees} (AST) from textual descriptions. 
The current state-of-the-art, \textsc{SketchAdapt}~\cite{nye2019learning}, uses a combination of neural and sketch-based approaches for program synthesis. As a downside of neural modeling, both paradigms necessitate large volumes of datasets in fully supervised settings. However, due to the rare availability of good quality and large-scale real datasets, these approaches leverage synthetic datasets. Popular synthetic datasets include \textsc{AlgoLisp}~\cite{neuralprogramsearch2018}, NAPS~\cite{zavershynskyi2018naps}, Karel~\cite{karel}, WikiSQL~\cite{sql} and dataset of bash commands~\cite{LinWPVZE2017:TR}. In this paper, we extensively experiment with \textsc{AlgoLisp} dataset under PBD settings ($\mathbb{I}$$\rightarrow$$\mathbb{O}$, see Figure~\ref{fig:example_problem}). 

\noindent\textbf{\underline{Robustness} Against Adversarial Attacks}
Recently, we witness a growing interest in evaluating deep learning models against adversarial attacks~\cite{yuan2019adversarial}. However, to the best of our knowledge, we do not find any work that evaluates the adversarial robustness of neural program synthesis systems. Specifically, we are interested in answering questions like \textit{``Are generative models trained on synthetically constructed datasets sufficiently robust against adversarial attacks?''}  In this paper, we evaluate automatic program synthesis models trained on synthetic datasets against adversarial attacks. We propose different classes of adversarial attacks and show the inability of state-of-the-art code generation models to generalize to extremely elementary test examples.  

\noindent\textbf{\underline{Self-supervised} Paradigm For Automatic Code Synthesis}
The field of Natural Language Processing (NLP) is witnessing a significant paradigm shift towards self-supervised learning.
Thus, some of the pertinent questions from the current context can be \textit{``Can self-supervised learning paradigm provide near state-of-the-art program synthesis performance?''} We experiment with several variants of  Transformer-based encoder-decoder models to showcase that self-supervised models significantly outperform previous neural approaches like RNN and LSTM based attention architectures~\cite{bednarek2018ain,nye2019learning,neuralprogramsearch2018}. Besides, to understand the adversarial robustness, we answer questions like \textit{``Are self-supervised models robust against adversarial attacks?''}
We show that the transformer architecture uses relatively lesser training text (only problem description and no I/O pairs) and is more robust against adversarial attacks than traditional neural code generation models. 

\noindent\textbf{\underline{Debiasing} The Synthetic Datasets}
Towards the end, we conduct extensive experiments to showcase the role of the synthetic datasets in introducing training bias and propose simple dataset augmentation strategies to debias the synthetic datasets. Extensive experimentation shows an increase in the robustness against adversarial attacks. 

\noindent \paragraph{Key Contributions:} 
To summarize, the key contributions of our work are:
\begin{itemize}[nosep,noitemsep]
    \item Identification and formalization of adversarial attacks on the existing program synthesis tools as well as an evaluation of these tools on a test suite of  manually-generated adversarial examples; 
    \item A new automatic code synthesis system, \textsc{AutoCoder}, and its robust comparison against state-of-the-art neural code generation systems; and 
    \item A generative algorithm to automatically augment the original AlgoLISP dataset with a large number of adversarial examples.
\end{itemize}

\section{Problem Definition}
 After introducing the general program synthesis paradigm in the previous section, we are now in a position to define the DSL-based program synthesis problem formally. Given a DSL $L$, we aim to learn a synthesis algorithm $A$ such that given a set of text description and its corresponding code snippet, {($i_1$,$o_1$), $\dots$, ($i_n$,$o_n$)}. The synthesis algorithm $A$ learns a program $P\in L$, such that it satisfies all the corresponding input-output test cases $e_j$'s of description ($NL$) and code snippet ($i_j$,$o_j$) pair, i.e., 
\begin{equation}
\begin{aligned}
\label{eq:problem}
    \forall{j,k}: P(e_{j(k,in)}) = e_{j(k,out)}: \\
    \;1\le j \le n \;\&\; 1\le k\le l
\end{aligned}
\end{equation}

Where, $e_{j(k,in)}$ and $e_{j(k,out)}$ represents the input and output of the  $k^{th}$ test case of description and code snippet ($i_j$,$o_j$) pair, respectively. Here, $l$ represents the number of test cases corresponding to each description and code snippet pair. Note that, in Eq.~\ref{eq:problem}, we match the test cases and not the actual generated code; a given textual description can possibly generate structurally dissimilar variants of the ground truth code, preserving the logical functionalities.

Formally, an adversarial text description $(NL')$ for a program synthesis model generates a program ($P_{adv}$) such that:
\begin{equation}
\begin{aligned}
\label{eq:problem1}
    \forall{j,k}: P_{adv}(e_{j(k,in)}) \ne e_{j(k,out)}: \\
    \;1\le j \le n \;\&\; 1\le k\le l
\end{aligned}
\end{equation}

under the constraint that-
\[||NL'-NL|| \le \delta\]

\noindent where $\delta$ denotes the amount of perturbation. Let $P_{orig}$ denotes the program corresponding to $NL$ and $P_{adv\_sol}$ corresponds to a program that can correctly solve $NL'$. Depending on whether $P_{adv\_sol}$ is the same as $P_{orig}$, attacks can be classified into the following two categories:

\textbf{Program Invariance Attacks:} In these types of attacks, we perturb $NL$ such that the original program is also a solution of $NL'$ i.e., ($P_{orig}=P_{adv\_sol}$).

\textbf{Program Directional Attacks:} In these type of attacks, we perturb $NL$ such that the  original program is not a solution of $NL'$ i.e., ($P_{orig} \ne P_{adv\_sol}$).  

\section{Dataset}
\label{sec:datasets}
In this paper, we use a synthetically constructed code generation dataset \textsc{AlgoLisp}~\cite{neuralprogramsearch2018}. \textsc{AlgoLisp} is constructed over a domain-specific language (DSL), inspired by Lisp. Instead of existing programming languages (like Python, C, or Java), DSLs provide flexibility in converting to other target languages and adding constraints to simplify its automated generation~\cite{neuralprogramsearch2018}. The dataset comprises the problem description and the corresponding implementations of the problem in a Lisp-inspired DSL. Each problem description is accompanied by a code snippet and 10 test cases. Each test case is an input/output pair, where input is to be fed into the synthesized program, and output represents the expected output the program should produce. Figure~\ref{fig:example_problem} illustrates an example problem showing a textual description, its corresponding Lisp DSL program tree, and few I/O test pairs. Overall, the dataset contains 100,000 problems with average textual description length and average code snippet length of 37.97 and 45.13 characters. The \textsc{AlgoLisp} dataset comprises train, validation, and test split of 79214,  9352, and 10940 examples, respectively. The average depth of the program tree is 10.28. Table~\ref{table:dataset} lists the detailed statistics of original \textsc{AlgoLisp} dataset. 

\begin{table}[]
    \centering
    \begin{tabular}{l|c|c}
    \toprule
        & \textbf{Original} & \textbf{Filtered}\\\hline
        No. of instances  & 100,000 & 90,153 \\ 
        Avg. text length  & 37.97 & 37.75\\ 
        Avg. code depth & 10.35& 10.28\\ 
        Avg. code length & 45.13& 44.86\\ 
        Vocabulary size & 288 & 287\\
        \bottomrule
    \end{tabular}
    \caption{Statistics of the \textsc{AlgoLisp} dataset.}\label{table:dataset}
\end{table}

In 2018, Bednarek~\etal~\cite{bednarek2018ain} showed multiple instances of compilation errors in the original \textsc{AlgoLisp} dataset. Specifically, the DSL compiler fails to pass I/O pairs with ground truth code. They, therefore, constructed a filtered subset of \textsc{AlgoLisp} dataset containing only those problem instances that pass all the input-output test cases\footnote{Even though we find few instances that resulted in partial passing of test cases.}. Overall, the filtered dataset contains 90,153 instances. Table~\ref{table:dataset} also details the  statistics of the filtered \textsc{AlgoLisp} dataset. 
To the best of our knowledge, except NAPS~\cite{zavershynskyi2018naps} and Karel~\cite{karel}, no similar code synthesis dataset exists that contains problem description along with the test cases and other meta information.  Popular datasets like JAVA~\cite{java}, WikiSQL~\cite{sql} only contain problem description and the corresponding code, leading to limitations in evaluating structurally different but logically similar synthesized codes. Although NAPS and Karel contains all the required meta-information, Karel does not deal with any natural language; it is a robotic programming language. On the other hand, in our internal data analysis, NAPS shows several data inconsistencies\footnote{The NAPS dataset is very noisy due to crowd-sourcing.} such as the presence of a long sequence of characters like {\ttfamily{abcdabcd}}, which conveys no meaning and inconsistent tokenization of sentences. 

\section{The SOTA Code Generation Models}
\label{sec:sota_works}
In this paper, we thoroughly experiment with state-of-the-art DSL-based code generation model, \textbf{SketchAdapt}~\cite{nye2019learning}. SketchAdapt (hereafter \textit{`SA'}) synthesizes programs from textual descriptions as well as input-output test examples. It combines neural networks and symbolic synthesis by learning an intermediate `sketch' representation. It has been demonstrated empirically that \textsc{SA} recognizes similar descriptive patterns as effectively as pure RNN approaches while matching or exceeding the generalization of symbolic synthesis methods. The \textsc{SA} system consists of two main modules: 1) a sketch generator and 2) a program synthesizer. Given an encoded input description, the sketch generator generates a distribution over program sketches. The generator is trained using a sequence-to-sequence recurrent neural network with attention to assign a high probability to sketches that are likely to yield programs satisfying the specification. The program synthesizer takes a sketch as a starting point and performs an explicit symbolic search to ``fill in the holes'' in order to find a program that satisfies the specification.
The pre-trained model, along with the relevant codebase, is available at \url{https://github.com/mtensor/neural_sketch}.

Additionally, we found two more relevant baselines, Structure-Aware Program Synthesis ~\cite{bednarek2018ain} and \textsc{Seq2Tree} model~\cite{Ma2017Seq2TreeAT}. Structure-Aware Program Synthesis (hereafter, \textit{`SAPS'}) adapts the program synthesis problem under the Neural Machine Translation framework by employing a bi-directional multi-layer LSTM network for generating code sequence corresponding to textual descriptions. The Seq2Tree model consists of a sequence encoder and a tree decoder. The sequence encoder reads the problem description, and a tree decoder augmented with attention computes probabilities of each symbol in a syntax tree node one node at a time. However, both baselines cannot be implemented due to the unavailability of a code repository or the pre-trained model\footnote{The results cannot be reproduced due to missing experimental details.}. We, therefore, thoroughly experiment with \textit{SA} as the only baseline system.

\noindent \textbf{Evaluating Generation Performance:}
We evaluate the above state-of-the-art code synthesis systems on the filtered \textsc{AlgoLisp} dataset.  We, verbatim, follow the experimental settings presented in SA and \textsc{SAPS}.  Note that, in the filtered dataset, the number of test cases is lesser than the original dataset. 
\begin{table}[!t]
\centering
\begin{tabular}{lc}
\toprule

\bfseries Model&\bfseries Accuracy Scores\\\hline
\textsc{SA}   &0.958\\
\textsc{SAPS}*   & 0.929\\
\textsc{Seq2Tree}* & $0.858^\dagger$\\
\hline
\textsc{VAC} &\textbf{0.968} \\
\textsc{GAC} &0.963 \\

\bottomrule
\end{tabular}
\caption{Comparing state-of-the-art code generation models. * represents accuracy scores taken, verbatim, from the corresponding papers due to unavailability of code or pretrained model. $^\dagger$ represents accuracy scores on the original test set.}\label{tab:accuracy}
\end{table}
At the same time, the training data remains the same as the original dataset. We compute accuracy scores ($A$) for performance evaluation on holdout test set defined as $A = \frac{n}{N}$, where $n$ is the number of problems for which the generated code passes all the 10 test cases and $N$ is the total number of problems in the holdout test set. Table~\ref{tab:accuracy} shows the accuracy scores for three state-of-the-art code generation systems. As expected, \textsc{SA} outperformed the rest of the two baseline systems with a significant margin. 

\section{\textsc{AutoCoder}}

\begin{figure*}[!t]
    \centering 
    \includegraphics[width=0.9\linewidth]{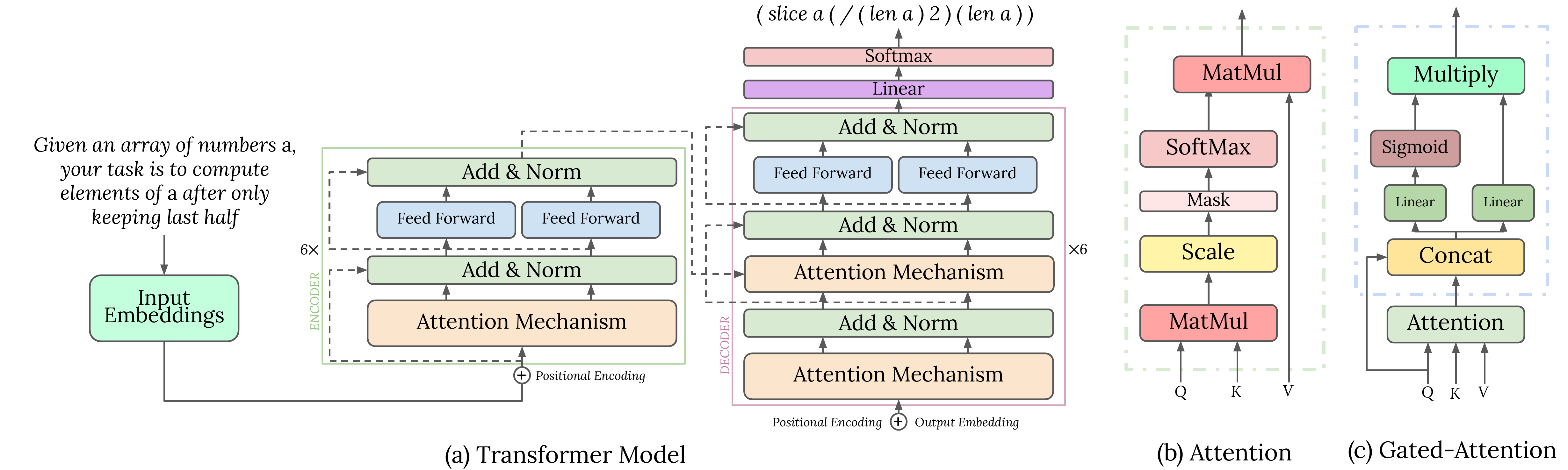}
\caption{Transformer model. (b) The self-attention mechanism. (c) The gated-attention mechanism. 
}
\label{fig:model}

\end{figure*}

In this section, we discuss our implementation of the neural model \textsc{AutoCoder} to address the automatic code generation problem. Recently, Transformers~\cite{DBLP:journals/corr/VaswaniSPUJGKP17} have shown state-of-the-art performance for several Natural Language Processing tasks~\cite{wang2019learning,li2019neural,abzianidze2019first}, including machine translation, classification, etc. Inspired by its success, we propose a transformer-based code-generation model to generate code based on natural language descriptions automatically. Specifically, the model encodes the textual description using multiple layers of encoders and then decodes a program token-by-token while attending to the description. The basic pipeline of our proposed model is analogous to the simpler sequence-to-sequence models that are employed for similar generation tasks. These models usually have an encoder-decoder architecture~\cite{cho2014learning,britz2017massive}. The encoder maps an input sequence of symbol representations consisting of tokens $x_1,x_2,\cdots,x_n$ to intermediate representation which the decoder then generates an output sequence $y_1,y_2,\cdots,y_m$ of symbols one element at a time. At each step, the model is auto-regressive, consuming the previously generated symbols as additional input when generating the next output symbol. As depicted in Figure~\ref{fig:model}, we utilize the core Transformer~\cite{DBLP:journals/corr/VaswaniSPUJGKP17} model implementation and propose significant structural alterations in the attention module to develop \textsc{AutoCoder}.

\noindent\textbf{The encoding and decoding layers:}  We keep the number of encoder layers (=6) the same as the core Transformer model. Similar to core implementation, we keep the output dimension size as 512 in all the encoding sub-layers of the model as well as in the embedding layers. The decoder side is also stacked with six identical layers. In Figure~\ref{fig:model}a, the sub-layers of encoder and decoder layers are described using standard notations. 

\noindent\textbf{The different attention mechanisms:} Attention mechanisms have become an integral part of sequence modeling tasks, allowing modeling of dependencies without regard to their distance in the input or output sequences. Specifically, we experiment with two attention mechanisms: (i) vanilla self-cross attention and (ii) gated cross attention. The Vanilla self-cross attention is the basic attention mechanism used in traditional transformer models~\cite{DBLP:journals/corr/VaswaniSPUJGKP17}. The self-attention module relates different positions of an input sequence in order to compute a representation of the input sequence. This module is present in both encoder and decoder layers. The cross attention module relates different positions of an input sequence to the output sequence. This module connects the encoder and decoder components. We term \textsc{AutoCoder} variant that uses the standard self-cross attention module as \textbf{\textsc{Vanilla-AutoCoder} (VAC)}. 
\begin{equation}
    f_{SA}=f_{dot}(Q_e,K_e,V_e)=softmax\left( \frac{Q_eK_e^T}{\sqrt{d}} \right) V_e
\end{equation}

\begin{equation}
    f_{CA}=f_{dot}(Q_e,K_d,V_d)=softmax\left( \frac{Q_eK_d^T}{\sqrt{d}} \right) V_d
\end{equation}
where $f_{dot}$ represents scaled dot product attention~\cite{DBLP:journals/corr/VaswaniSPUJGKP17}, ($Q_e,K_e,V_e$) denotes encoder sequence representations in terms of query, key and value respectively, and ($Q_d,K_d,V_d$)  denotes decoder sequence representation in terms of query, key and value respectively. 

The gated cross attention mechanism filters out the unnecessary part of the sequence by attending over the generated cross attention scores $f_{CA}$ and determining the relevant attention. The gated cross attention module ($f_{GA}$) uses a sigmoidal gating mechanism for filtering out irrelevant cross attention while decoding the output. It generates an \textit{information vector}($i$) which carries relevant representation of the input vector and an \textit{attention gate} ($g$) that filters the relevant attention scores. Now this filtered attention is applied to the information vector to obtain \textit{attended information}, 
or the relevant information.
\begin{equation}
\begin{aligned}
 f_{GA}=\sigma(W_q^gQ_e + W_v^gf_{CA} + b^g) \\ 
 \odot (W_q^iQ_e + W_v^if_{CA} + b^i)
\end{aligned}
\end{equation}

$\sigma$ denotes sigmoid activation, $\odot$ denotes element-wise product, $W_q^i$ and $W_v^i$ represent weight matrices corresponding to value query and value at \textit{information vector}, respectively,   $W_q^g$ and $W_v^g$ represent weight matrices corresponding to value query and value at \textit{attention gate}, respectively. Note that, $\{W_q^i,W_v^i,W_q^g,W_v^g\} \in \mathbb{R}^{d \times d}$ and $\{b^i,b^g\} \in \mathbb{R}^{d}$. Figure~\ref{fig:model}c shows the gated cross-attention architecture. We term \textsc{AutoCoder} variant that uses gated cross attention as \textbf{\textsc{Gated-AutoCoder} (GAC)}.

\noindent \textbf{Comparing \textsc{AutoCoder} against baselines:} Table~\ref{tab:accuracy} also compares \textsc{AutoCoder} variants against baseline systems. Both variants outperformed the three baselines. Among the two variants, VAC performed marginally better than GAC. To summarize, the results showcase that even simple Transformer variants can result in high gains in code synthesis. 

\section{The Adversarial Experiments}
\label{sec:adversarial}

\subsection{Adversarial Attack Types}
We define five classes of adversarial examples. All our proposed attacks are black-box un-targeted attacks. Our attacks do not have any knowledge of the target model, nor does it have any information about the gradients and model parameters. Table~\ref{tab:adversarial_examples} shows representative examples of actual descriptions and corresponding adversarial descriptions. The classes are:  
\begin{enumerate}[nosep,noitemsep]
    \item \textbf{Variable Change (VC):} Changing single and multi-character variables and argument names in the original problem description, input arguments, and program trees to examine if the model correctly generates the corresponding output code. 
    \item \textbf{Redundancy Removal (RR):} Removing filler or redundant words without affecting the overall meaning of the input description. 
    \item \textbf{Synonym Replacement (SR):} Replacing words with their corresponding synonyms. 
    \item \textbf{Voice Conversion (VoC):}  Converting a problem description in the active voice to its corresponding passive voice. 
    \item \textbf{Variable Interchange (VI):} Interchanging variable names in problem descriptions comprising multiple variables. 
\end{enumerate}

The classes \textbf{VI} and \textbf{VC} belong to program directional attack category, whereas classes \textbf{RR}, \textbf{SR}, \textbf{VoC} belong to program invariance attack category. For example, consider the representative example for \textbf{VC} class in Table \ref{tab:adversarial_examples}, changing variable name from {\ttfamily{a}} to {\ttfamily{b}} led to the change in the ground truth program that can solve the problem i.e. from {\ttfamily{(strlen a)}} to {\ttfamily{(strlen b)}}. Now, model predicting any other token except the variable {\ttfamily{b}} is an adversary.  In case of \textbf{RR}, removing redundant token is a program invariance perturbation, hence the ground truth program remains unchanged.

\begin{table}[!t]
    \centering
    \small{
    \begin{tabular}{c|p{11.2 cm}}
    \toprule
    \textbf{Class}&\textbf{Representative Example}\\\hline
        \multirow{4}{*}{VC} & \textbf{OD:}  \textcolor{blue}{Given a string a, what is the length of a.} \\
        &\textbf{OO:} 
        \textcolor{blue}{\small{\ttfamily{(strlen a)}}}\\ 
        & \textbf{AD:} \textcolor{mauve}{Given a string b, what is the length of b.} \\
         & \textbf{AO:} \textcolor{mauve}{\small{\ttfamily{(strlen a)}}}\\\hline
        
        \multirow{7}{*}{RR} & \textbf{OD:} \textcolor{blue}{Given a number a, compute the product of \textbf{all} the numbers from 1 to a.} \\
        &\textbf{OO:} 
        \textcolor{blue}{\small{\ttfamily{(invoke1 (lambda1 (if ( $\leq$ arg1 1 )1(*( self( -arg1 1 )) arg1 ))) a)}}}\\ 
        & \textbf{AD:} \textcolor{mauve}{Given a number a, compute the product of the numbers from 1 to a.}  \\
       
         & \textbf{AO:}  \textcolor{mauve}{\small{\ttfamily{( * a 1 )}}}\\ \hline

         \multirow{8}{*}{SR} & \textbf{OD:}  \textcolor{blue}{consider an array of numbers, what is reverse of elements in the given array that are odd} \\
        &\textbf{OO}: 
        \textcolor{blue}{\small{\ttfamily{(reverse ( filter a ( lambda1 ( == ( \% arg1 2 )1))))}}}\\ 
        & \textbf{AD:} \textcolor{mauve}{consider an array of numbers, what equals reverse of elements in the given array that are odd} \\
         & \textbf{AO:} \textcolor{mauve}{\small{\ttfamily{(reduce ( filter a ( lambda1 ( == ( \% arg1 2 )1))))}}}\\\hline

         \multirow{7}{*}{VoC} & \textbf{OD:}  \textcolor{blue}{Given a number a, your task is to compute a factorial} \\
        &\textbf{OO}: 
        \textcolor{blue}{\small{\ttfamily{invoke1(lambda1(if(<= arg1 1) 1 (*(self(-arg1 1)) arg1)))a)}}}\\ 
        & \textbf{AD:} \textcolor{mauve}{Your task is to compute a factorial, given a number a} \\
         & \textbf{AO:} \textcolor{mauve}{\small{\ttfamily{(filter a ( partial1 b >))}}}\\ \hline

         \multirow{10}{*}{VI} & \textbf{OD:}  \textcolor{blue}{you are given an array of numbers a and numbers b, c and d, define e as elements in a starting at position b ending at the product of c and d ( 0 based ), what is e} \\
        &\textbf{OO}: 
        \textcolor{blue}{\small{\ttfamily{( slice a d ( * c b ) )}}}\\ 
        & \textbf{AD:} \textcolor{mauve}{you are given an array of numbers a and numbers b , c and e , define d as elements in a starting at position b ending at the product of c and e ( 0 based ) , what is d} \\
         & \textbf{AO:} \textcolor{mauve}{\small{\ttfamily{( slice a d ( * c b ) )}}}\\
         
    \bottomrule
    \end{tabular}}
    \caption{Representative examples from each adversarial class. Here, OD, OO, AD, and AO represent the original description, original output, adversarial description, and adversarial output, respectively.}
    \label{tab:adversarial_examples}

\end{table}

\subsection{Adversarial Performance}
\label{sec:adv_old}
In this section, we discuss the adversarial instance construction process. We construct adversarial examples using the holdout test instances following classwise constraints in a semi-supervised fashion. For example, an adversarial instance belonging to the \textbf{VI} class can only be generated if the problem description contains two or more variables. In addition, we used several NLP libraries for basic linguistic tasks. For example, we use the NLTK library to stochastically remove some stopwords from the program descriptions to generate instances for \textbf{RR} class. Similarly, we leverage POS tagging to identify active/passive voice to construct instances for the \textbf{VoC} class. And POS tagging and Wordnet hierarchies to construct instances for \textbf{SR} class.  Overall, we use about 1000 adversarial instances, equally divided per adversary class, for evaluating program synthesis systems. 

Table~\ref{tab:ad_test} presents generation performance of \textsc{SA} under adversarial settings using error percentage i.e. (100 - Accuracy \%), lower the error \% better is the adversarial robustness. Surprisingly, \textsc{SA} fails to generalize and produce significantly poor results under the adversarial setting. In particular, it performs very poorly on the \textbf{VoC} and variable \textbf{VI} classes. 
\begin{table}[!t]
    \centering
    \small{
    \begin{tabular}{c|ccc||ccc}
    
        \multirow{2}{*}{\textbf{Adv. Class}}  & \multicolumn{3}{c||}{\textbf{Error (\%)}} & \multicolumn{3}{c}{\textbf{Distance}}\\ 
        
        &\textbf{SA}& \textbf{VAC} &\textbf{GAC}& \textbf{Lev} &\textbf{LevR} & \textbf{BERT}\\\hline
        VC  & 48.0 & \textbf{42.5} & \textbf{42.5} & 2.24 & .05 & .005\\
        RR & 4.70 & 3.70 & \textbf{3.20} &4.55 & .13 & .044\\
        SR & \textbf{5.70} & 8.10 & 8.10 & 1 & .03 & .013\\
        VoC & 70.2  & 24.9& \textbf{24.4}& 16.54 & .54 & .015 \\
        VI & 70.0 & 67.7 &\textbf{67.2}  &  4.2& .08 & .043\\ 
        
   \end{tabular} 
   }
    \caption{Error percentage (columns 2--4) of SA, VAC and GAC for different adversarial classes. 
    Distance between (columns 5--7) adversarial and the corresponding original description. }
    \label{tab:ad_test}
\end{table}
This is because the model does not predict the correct code when the sentences that are generally active in the dataset are converted to passive sentences. Further, in our analysis, if variables {\ttfamily{b}} and {\ttfamily{d}} are interchanged, the model fails to recognize this change and outputs code as if no change has been done on the input sentences. Table~\ref{tab:ad_test} also presents generation performance of \textsc{AutoCoder} under adversarial settings.  \textsc{AutoCoder} variants show more robustness than \textsc{SA} in four out of five classes. We observe that one of the possible reason for the poor performance of \textsc{AutoCoder} variants is incorrect cross attending. For example, the variable {\ttfamily{a}} in the output is not attending the corresponding variable {\ttfamily{a}} in the problem description.

Even though \textsc{AutoCoder} showed more robustness than \textsc{SA} under adversarial settings, we observe a significant drop in the overall performance in both systems. We claim that the performance drop under the adversarial setting is attributed to bias in the synthetic dataset generation process. Some of the potential bias scenarios are: (1) small set of chosen variable names, (2) limited number of operations, (3) limited vocabulary usage, (4) variables occur in a sequential and alphabetical manner.

\subsection{Measuring Extent and Quality of Perturbations}

\subsubsection{Extent of Perturbations}
To measure the extent of perturbation in our proposed adversarial attacks, we experiment with the following two distance metrics: 

\textbf{Edit Distance:} We use the popular Levenshtein distance (hereafter, \textit{`Lev'}) to calculate the distance between adversarial description and the corresponding original description. It is defined as the minimum number of edit operations (delete, insert and substitute) required to convert one string to the other. We also report the ratio of Levenshtein distance to the length of sentences (hereafter, \textit{`LevR'}) to measure the extent of perturbation per length of the sentence. 
Table~\ref{tab:ad_test} (columns 5 and 6) shows distance values for the five adversarial classes. Except for \textbf{VoC} where the entire sentence structure changes, the other classes comprise examples constructed from significantly low perturbations. Note that, we limit the perturbation rate in \textbf{SR} to 1, as higher perturbations were leading to out-of-vocabulary problems and other grammatical inconsistencies. 

\textbf{Embedding Similarity:} We also measure the cosine similarity between adversarial description and the corresponding original description using sentence embeddings derived from pretrained model BERT~\cite{devlin2019bert}. The sentence embeddings are derived from a Siamese network trained using triplet loss~\cite{sent-bert}. We convert the similarity value into a distance value by subtracting it by 1 (hereafter, \textit{`BERT'}). We keep the embedding length as 768. 
Table~\ref{tab:ad_test} (column 7) reiterate the distance-based observations. Note that, as contextual embeddings successfully capture voice-related changes, the adversarial class \textbf{VoC} also shows low perturbation distance. 



\subsubsection{Human Evaluation}
We employ two undergraduate students expert in programming to evaluate the quality of constructed adversarial attacks. For this experiment, we randomly select ten instances from each adversary class along with the corresponding original instance (a sample dataset of a total of 100 instances). We first-of-all educate evaluators about the task by presenting them a set of program descriptions from the original ALGOLISP dataset. Next, we instruct them to evaluate each instance in the sampled set based on the following two criteria:

\textbf{Grammatical Correctness:} We ask the evaluators to rate the grammatical correctness of the sentences on a scale of 1--5. The rating of 1 being `completely grammatically incorrect description' and 5 representing `grammatically sound and correct'. 

\textbf{Naturalness:} We also ask the evaluators to judge the quality of the sentences on the basis of \textit{naturalness} of the texts i.e., how likely the test samples are drawn from the original data distribution. We ask to rate each sample on a scale of 1--5. The rating of 1 being `completely outside data distribution/unfamiliar example' and 5 representing `definitely from original data distribution'. 

\begin{table}[!t]
    \centering
    \small{
    \begin{tabular}{c|ccc|ccc}
    \toprule
         \multirow{2}{*}{\textbf{Adv. Class}}  & \multicolumn{3}{c|}{\textbf{Grammatical Score}} & \multicolumn{3}{c}{\textbf{Naturalness Score}}\\
         &Original&Adversarial&\%confusion&Original&Adversarial&\%confusion\\\hline
        
        VC   & 4.2 &\textbf{4.25} & 99\%& \textbf{3.95} &3.85& 98\%\\\hline
        
        RR & \textbf{4.20} &3.60& 88\% & \textbf{4.15} &3.60& 89\%\\ \hline
        
        SR & \textbf{4.40} & 3.85 & 90\% & \textbf{4.25}& 3.90& 92\%\\ \hline
        
        VoC   & \textbf{4.00}& 3.45 &89\%& \textbf{3.90} &3.65& 98\%\\ \hline
         
        VI  & \textbf{3.70} &3.50& 96\% &3.45 &\textbf{3.60}& 95\%\\\hline\hline
        
        \textbf{Average} &\textbf{4.10} &  3.73 & 92.4\% & \textbf{3.94}  & 3.71 & 94.4\%\\
        \bottomrule
   \end{tabular} 
   }
    \caption{Class-wise comparison of human evaluation results}
    \label{tab:human_result}
    
\end{table}


We summarize the human evaluation experiment in Table \ref{tab:human_result}. As evident from the table, the grammatical score and naturalness score of original sentences are higher than adversarial sentences. The evaluators were correctly able to identify the minor grammatical mistakes present in the \textbf{RR} class. Also, since changing the variables only does not add much human notable noise, evaluators were finding it difficult to distinguish between original sentences and adversarial sentences for \textbf{VC} and \textbf{VI} classes as depicted in the results of Table \ref{tab:human_result}. We also present the \% confusion score that reflects how much difficulty evaluators are facing in distinguishing between adversarial and original sentences.
Mathematically, it is defined as  $\%confusion = \Big( 1- \frac{|\textit{ original value - adversarial value }|}{5} \Big)\times 100$. The high \% confusion scores in  Table~\ref{tab:human_result} showcase the quality of constructed adversarial examples.  

\begin{table}[!b]
   
    \centering
    \small{%
    \begin{tabular}{c|c|p{11.2cm}}
    \toprule
         \multirow{2}{*}{\rotatebox[origin=c]{90}{\textbf{\centering RD}}}&\textbf{OS} & Consider an array of numbers a, your task is to find if a reads \textbf{the} same from both ends. \\
        &\textbf{FS} & Consider an array of numbers a, your task to find if a reads same from both ends.\\
        \hline
         \multirow{3}{*}{\rotatebox[origin=c]{90}{\textbf{\centering RI}}}&\textbf{OS} &Consider an array of numbers a, your task to find if a reads same from both ends.\\
        &\textbf{FS} & Consider \textbf{on} an array of \textbf{regular} numbers a, your task is to find if a reads the same from both ends\\
        
        \hline
         \multirow{2}{*}{\rotatebox[origin=c]{90}{\centering \textbf{RS}}}&\textbf{OS} & Consider an array of \textbf{numbers} a, your task is to find if a reads the \textbf{same} from both ends
        \\
        &\textbf{FS} & Consider an array of \textbf{integers} a, your task is to find if a reads the \textbf{integers} from both ends\\
        
        \hline
        \multirow{5}{*}{\rotatebox[origin=c]{90}{\textbf{ \centering BT}}}&\textbf{OS} & Given arrays of numbers a and b, what is the difference of elements of a and median in b. \\
        &\textbf{IS} & Was ist der Unterschied zwischen den Elementen von a und dem Median in b, wenn Arrays von Zahlen a und b gegeben sind? \\
        &\textbf{FS} & What is the difference between the elements of a and the median in b given arrays of numbers a and b?\\
        
        \hline
        \multirow{2}{*}{\rotatebox[origin=c]{90}{\centering \textbf{AR}}}&\textbf{OS} & you are given \textbf{an} array of numbers a, find not prime values in a
        \\
        &\textbf{FS} & you are given \textbf{at} array of numbers a, find not prime values in a\\
        
        \bottomrule
    \end{tabular}}
    \caption{Illustrative examples of different operations to modify \textsc{AlgoLisp} dataset. Here OS, IS and FS represents original, intermediate and final sentence, respectively.}
    \label{tab:example_BT}
\end{table}

\section{\textsc{AlgoLisp++}: The Debiased Dataset}

To mitigate the poor performance of \textsc{SA} and \textsc{AutoCoder} variants under adversarial settings, we extend the original \textsc{AlgoLisp} by adding a highly diversified collection of examples. However, as we see in previous sections, the automatic synthesis of instances is a challenging task. We, therefore, present a automatic instance generation algorithm inspired by the concepts of basic string editing~\cite{wei2019eda}, back translation~\cite{sennrich2016improving} and neural editing~\cite{hsieh-etal-2019-robustness}. We propose the following three classes of operations:

\begin{enumerate}[noitemsep,nosep]
\item \textbf{Basic editing operations (BE)}
We randomly edit tokens from the descriptions except few tokens that convey high semantic importance with respect to the programming languages. For example, the token {\ttfamily{concatenation}} conveys special meaning to the sentence and should not be edited. We reserve  $\sim10\%$ of the vocabulary tokens as non-editable. The non-editable list includes tokens such as {\ttfamily{times}}, {\ttfamily{sum}}, {\ttfamily{digits}}, {\ttfamily{maximum}}, {\ttfamily{prime}}, {\ttfamily{last}}, 
etc. We define a parameter $\alpha$ to regulate the number editable tokens in a sentence. The number of editable token is given by $\lfloor \alpha L \rfloor$, where $L$ is the length of the sentence. In our experiments, we assign $\alpha = 0.1$.  Next, we define three basic token-level edit operations: 

\noindent \textbf{Random Deletion (RD):} Randomly removing one or more words from the sentences.\\
\noindent \textbf{Random Insertion (RI): } Randomly inserting one or more words in the sentences.\\
\noindent \textbf{Random Substitution (RS):} Randomly substituting one or more words in the sentences. 

For RI and RS, we use BERT~\cite{devlin2019bert} uncased language model trained on monolingual English language. In RI, we randomly add $\lfloor \alpha L \rfloor$ masked tokens to the input sentence and predict tokens corresponding to these masked positions. In the case of RS, we randomly select $\lfloor \alpha L \rfloor$ editable tokens in a problem description and mask them. Further, the masked sentence is fed to the pre-trained BERT model to predict the mask tokens.   RD is reasonably straightforward as we randomly pick $\lfloor \alpha L \rfloor$ tokens from a sentence and delete them.

\item \textbf{Back-Translation (BT)}
In \textit{Back Translation (BT)}, a sentence is, first, translated to an intermediate language and again translated back to the original language. BT leads to the paraphrasing of the original sentence~\cite{sennrich2016improving}. In our case, the original language is English, and the intermediate language is German\footnote{We use German as one of the representative language due to the availability of good quality translations.}. Table~\ref{tab:example_BT} presents an illustrative example of a BT operation. We leverage native Google Translate API for English to German translation and vice-versa.

\item \textbf{Attention-based replace operation (AR)}
Inspired by the quality of augmented sentences in ~\cite{conaug}, we propose an attention-based augmentation operation that extracts the attention vector from the first encoder layer of the transformer and randomly replaces the maximally attended word with a random word in the vocabulary except the non-editable words to preserve the meaning of the sentence~\cite{hsieh-etal-2019-robustness}. 
\end{enumerate}

\begin{table}[!t]
 \centering
\small{
\begin{tabular}{l|ccc|ccc}
\multirow{2}{*}{\bfseries Name} & \multicolumn{3}{c|}{\bfseries Dataset Statistics}&\multicolumn{3}{c}{\bfseries Accuracy Scores}\\
&\bfseries Instances&\bfseries  Vocab. size &\bfseries Avg. length &\bfseries SA &\bfseries VAC & \bfseries GAC\\\hline
\textsc{AlgoLisp} & 79214 & 292  & 38.17 &0.958&\textbf{0.968}&0.963\\
\textsc{AlgoLisp++}& 142644 & 3152 &37.97&0.944&0.943&\textbf{0.947}\\
\bottomrule
\end{tabular}}
\caption{Statistics (columns 2--4) of \textsc{AlgoLisp} and \textsc{AlgoLisp++} training datasets. Accuracy scores (columns 5--7) of SA, VAC and, GAC for two \textsc{AlgoLisp} variants.}\label{tab:algolispspecs}
\end{table}

\begin{table}[!t]
    \centering
    \begin{tabular}{lccc}
    
            \multirow{2}{*}{\textbf{Adv. Class}}  & \multicolumn{3}{c}{\textbf{Error (\%)}}\\\cline{2-4}
        & \textbf{SA}& \textbf{\textsc{VAC}} & \textbf{\textsc{GAC}}\\\hline
        VC & {41.5}(48)& 36.00(42)& \textbf{34.40}(42)\\
        RR & 3.70(5) &\textbf{3.20}(4) & 4.20(5)\\
        SR & \textbf{4.40}(6) & 4.70(9) & 8.20(9)\\
        VoC &  \textbf{19.60}(71) & 24.40(25) & 23.60(24)\\
        VI & 67.90(70) & \textbf{62.50}(68) & 67.70(69)\\\hline
        \textbf{Average} & 27.4 & \textbf{26.1} & 27.1\\
        \bottomrule
    \end{tabular}
    \caption{Comparing adversarial robustness  of \textsc{AutoCoder} variants against \textsc{SA} for \textsc{AlgoLisp++}. The value present inside the bracket represent corresponding \textsc{AlgoLisp} error percentage.}
    \label{tab:final_result}
\end{table}

\noindent \textbf{The Generation Algorithm}
Algorithm~\ref{algo} details the data augmentation pipeline. Each sentence in the \textsc{AlgoLisp} dataset undergoes a series of edit operations parameterized by six free parameters $\rho_1$, $\rho_2$, $\rho_3$, $\sigma_1$, $\sigma_2$, and $\sigma_3$. $\rho_1$, $\rho_2$, and $\rho_3$ represent probability of token-level edit operations, back translation, and attention-based replace, respectively. $\sigma_1$, $\sigma_2$, and $\sigma_3$ represent  probability of deletion, insertion, and substitution, such that $\sigma_1 + \sigma_2 + \sigma_3 =1$.  In our experiments, we keep $\rho_1=0.5$, $\rho_2=0.2$ and $\rho_3=0.1$. In case of a length of a sentence greater than the average length, we assign $\sigma_1 =0.5$,  $\sigma_2=0.25$, and $\sigma_3=0.25$. Whereas if the length of a sentence lesser than the average length, we assign $\sigma_1 =0.2$,  $\sigma_2=0.4$, and $\sigma_3=0.4$. Overall, the augmentation approach has resulted in new 89,214 instances.
\begin{algorithm}[!t]
\caption{Generating \textsc{AlgoLisp++}.}\label{algo}

\begin{algorithmic}[1]
\small{
\REQUIRE $D \gets$ \textsc{AlgoLisp} dataset\\
\hspace{2em} $D' \gets$ \textsc{AlgoLisp++} dataset (initially empty) \\
\hspace{2em}$BE()$ $\gets$ performs basic edits \\
\hspace{2em}$BT()$ $\gets$ performs back translation \\
\hspace{2em} $AR()$ $\gets$ performs attention-based replace\\

\STATE $\Sigma=(\sigma_1,\sigma_2,\sigma_3)$
\FOR{each sample $\chi$ in $D$}
    \STATE toss coin with head prob. $\rho_1$
    \IF{head}
        \IF{LEN($ \chi $) $>$ AVG\_LEN(D)}
            \STATE Assign $\sigma_1 \ge \sigma_2$ and $\sigma_1 \ge \sigma_3$
        \ELSE
            \STATE Assign $\sigma_1 \le \sigma_2$ and $\sigma_1 \le \sigma_3$
        \ENDIF
        \STATE op $\gets$ sample an operation according to the multinomial distribution $\Sigma$
        \STATE Add BE($\chi$,op) to $D'$
    \ENDIF
    \STATE toss coin with head prob. $\rho_2$    
    \IF{head}
        \STATE Add BT($\chi$) to $D'$
    \ENDIF
\ENDFOR
\FOR{each sample $\chi$' in $\{D - D'\}$}
    \STATE toss coin with head prob. $\rho_3$ 
    \IF{head}
        \STATE Add AR($\chi$) to $D'$
    \ENDIF
    
\ENDFOR
\STATE Add all the examples of $D$ to $D'$

}
\end{algorithmic}\label{algo}
\end{algorithm}
Table~\ref{tab:algolispspecs} compares statistics of the newly constructed \textsc{AlgoLisp++} dataset against the  original \textsc{AlgoLisp} dataset.

\noindent \textbf{System evaluations on \textsc{AlgoLisp++}:}
Table~\ref{tab:algolispspecs} also compares code generation performance of \textsc{AutoCoder} variants against state-of-the-art system \textsc{SA} on \textsc{AlgoLisp++} dataset. 
We observe an overall marginal decrease in the generative performance of all systems under adversarial conditions. 
However, \textsc{AlgoLisp++} has resulted into high gains under adversarial setting (see Table~\ref{tab:final_result}). Specifically, \textsc{SA} shows more performance gain than \textsc{AutoCoder} variants under adversarial settings, especially in the VoC class with a decrease of more than 50 points in error percentage. 



\section{Conclusion}
\label{sec:conc}
In this paper, we propose a series of adversarial attacks to showcase limitations in SOTA code synthesis models' robustness. We experimented with Transformer-based model variants to showcase performance gain over previous SOTA systems and robustness under adversarial setting. Finally, we proposed a data augmentation pipeline to increase the adversarial robustness of code generation models.
In the future, we plan to extend our methodology and develop a general framework to study the adversarial robustness of code generation systems trained on synthetic and natural programming datasets.

\bibliography{ms}
\bibliographystyle{splncs04}

\end{document}